\documentclass[10pt,twocolumn,letterpaper]{article}

\usepackage{cvpr}
\usepackage{times}
\usepackage{epsfig}
\usepackage{graphicx}
\usepackage{amsmath}
\usepackage{amssymb}

\usepackage{url}
\usepackage{xcolor}
\usepackage{multirow}
\usepackage{amsmath}
\usepackage{array}
\usepackage{adjustbox}
\usepackage{caption,booktabs}
\usepackage{amssymb}
\usepackage{latexsym}

\usepackage[ruled, lined, commentsnumbered, longend]{algorithm2e}
\usepackage{xcolor}

\usepackage[pagebackref=true,breaklinks=true,letterpaper=true,colorlinks,bookmarks=false]{hyperref}

\cvprfinalcopy 

\def\cvprPaperID{****} 

\ifcvprfinal\fi
\begin{document}

\title{Contrastive Domain Adaptation}

\author{Mamatha Thota\\
University of Lincoln\\
Lincoln, LN6 7TS, UK\\
{\tt\small mthota@lincoln.ac.uk}
\and
Georgios Leontidis\\
University of Aberdeen\\
Aberdeen, AB24 3UE, UK\\
{\tt\small georgios.leontidis@abdn.ac.uk}
}

\maketitle

\begin{abstract}
Recently, contrastive self-supervised learning has become a key component for learning visual representations across many computer vision tasks and benchmarks. However, contrastive learning in the context of domain adaptation remains largely underexplored. In this paper, we propose to extend contrastive learning to a new domain adaptation setting, a particular situation occurring where the similarity is learned and deployed on samples following different probability distributions without access to labels. Contrastive learning learns by comparing and contrasting positive and negative pairs of samples in an unsupervised setting without access to source and target labels. We have developed a variation of a recently proposed contrastive learning framework that helps tackle the domain adaptation problem, further identifying and removing possible negatives similar to the anchor to mitigate the effects of false negatives.
Extensive experiments demonstrate that the proposed method adapts well, and improves the performance on the downstream domain adaptation task.
\end{abstract}

\section{Introduction}

Over the last few years, Deep Learning (DL) \cite{lecun2015deep} has been successfully applied across numerous applications and domains due to the availability of large amounts of labeled data, such as computer vision and image processing \cite{ribeiro2019deep, wu2020recent, thota2020multi, NEURIPS2020_47fd3c87}, signal processing \cite{caliva2018deep, purwins2019deep, he2020new}, autonomous driving \cite{liu2020video, wang2019deep, ghosh2019segfast}, agri-food technologies \cite{alhnaity2020autoencoder, khan2020iot}, medical imaging \cite{karakanis2020lightweight, li2020anatomical}, etc. Most of the applications of DL techniques, such as the aforementioned ones, refer to supervised learning, it requires manually labeling a dataset, which is a very time consuming, cumbersome and expensive process that has led to the widespread use of certain datasets, e.g. ImageNet, for model pre-training. On the other hand, unlabeled data is being generated in abundance through sensor networks, vision systems, satellites, etc. One way to make use of this huge amount of unlabeled data is to get supervision from the data itself. Since unlabeled data are largely available and are less prone to labeling bias issues, they tend to provide visual information independent from specific domain styles.

Nowadays, self-supervised visual representation learning has been largely closing the gap with, in some cases, even surpassing supervised learning methods. One of the most prominent self-supervised visual representation learning techniques that has been gaining popularity is contrastive learning, which aims to learn an embedding space by contrasting semantically positive and negative pairs of samples \cite{chen2020simple, chen2020big, he2020momentum}.

However, whether these self-supervised visual representation learning techniques can be efficiently applied for domain adaptation has not yet been satisfactorily explored. When, one applies a well performing model learned from a source training set to a different but related target test set, generally the assumption is that both these sets of data are drawn from the same distributions. When this assumption is violated, the DL model trained on the source domain data will not generalize well on the target domain, due to the distribution differences between the source and the target domains known as domain shift. Learning a discriminative model in the presence of domain shift between source and target datasets is known as Domain Adaptation.

Existing domain adaptation methods rely on rich prior knowledge about the source data labels, which greatly limits their application, as explained above. This paper introduces a contrastive learning based domain adaptation approach that requires no prior knowledge of the label sets. The assumption is that both the source and target datasets share the same labels, but only the marginal probability distributions differ. 

One of the fundamental problems with contrastive self-supervised learning is the presence of potential false negatives that need to be identified and eliminated; but without labels, this problem is rather difficult to solve. Some notable work related to this area has been proposed in \cite{kalantidis2020hard} and \cite{robinson2020contrastive}, where both methods focused on mining hard negatives; \cite{huynh2020boosting} developed a method for false negative elimination and false negative attraction and \cite{chuang2020debiased} proposed a method to correct the sampling bias of negative samples.

Over the past few years, ImageNet pre-training has become a standard practice, but using contrastive learning has demonstrated a competitive performance without access to labeled data by training the encoder using the input data itself. In this paper, we extend contrastive learning also referred as unsupervised representation learning without access to labeled data or pretrained imagenet weights, where we leverage the vast amount of unlabeled source and target data to train an encoder using random initialized parameters to the domain adaptation setting, a particular situation occurring where the similarity is learned and deployed on samples following different probability distributions. We also present an approach to address one of the fundamental problems of contrastive representation learning, i.e. identifying and removing the potential false negatives. We performed various experiments and tested our proposed model and its variants on several benchmarks that focus on the downstream domain adaptation task, demonstrating a competitive performance against baseline methods, albeit not using any source or target labeled data.

The rest of the paper is laid out as follows: Section 2 presents the related work in self-supervised contrastive representation learning and domain adaptation methods. Section 3 describes our proposed approach, Section 4 presents the datasets and experimental results on domain adaptation after applying our model, and finally, Section 5 summarizes our work and future directions.

\subsection{Contributions}
The main contributions of this work can be summarised as follows:
\begin{itemize}
    \item We explore contrastive learning in the context of Domain Adaptation, attempting to maximize generalization between source and target domains with different distributions. 
    
    \item We propose a Domain Adaptation approach that does not make use of any labeled data or involves imagenet pretraining.
    
    
    \item We incorporate false negative elimination to the domain adaptation setting, resulting in improved accuracy and without incurring any additional computational overhead.
    
    \item We extend our domain adaptation framework and perform various experiments to learn from more than two views. 
    
\end{itemize}

\begin{figure*}[t!]
\centering
\includegraphics[scale=0.47]{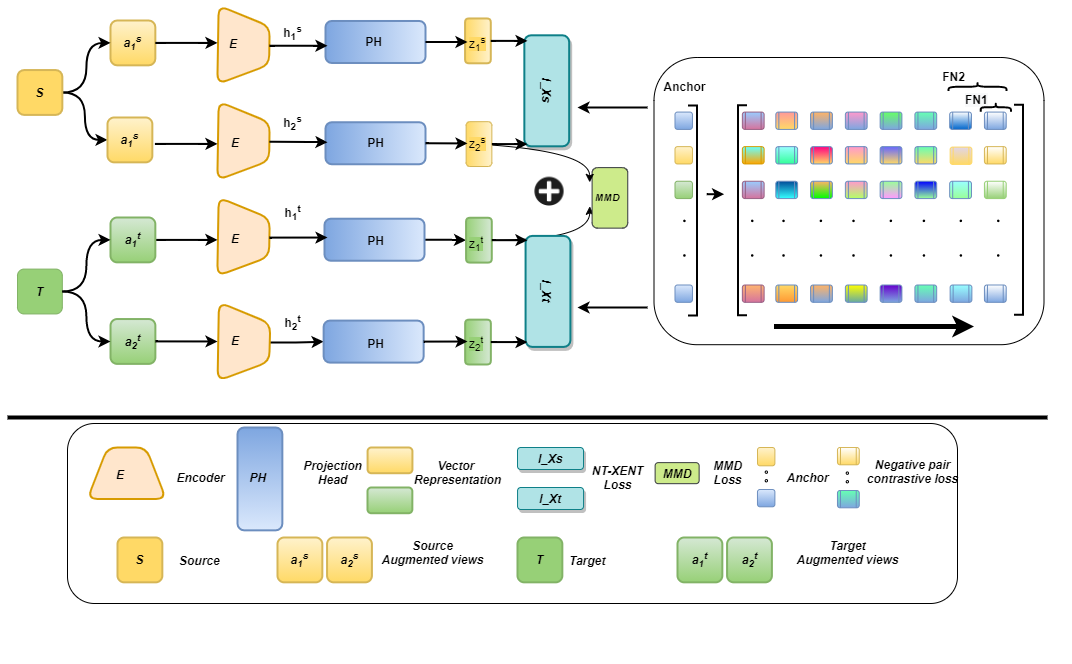}
\caption{Overview of our proposed Contrastive Domain Adaptation model. Image on the \textbf{Left}, shows the pipeline of our model and image on the \textbf{Right} shows the loss function. }
\label{fig-Network}
\end{figure*}

\section{Related Work}

\textit{Domain Adaptation}: Domain adaptation is a special case of transfer learning where the goal is to learn a discriminative model in the presence of domain shift between source and target datasets. Various methods have been introduced to minimize the domain discrepancy in order to learn domain-invariant features. Some involve adversarial methods like DANN \cite{ganin2016domain}, ADDA\cite{tzeng2017adversarial} that help align source and target distributions. Other methods propose aligning distributions through minimizing divergence using popular methods like maximum mean discrepancy \cite{gretton2012kernel, long2015learning, long2017deep}, correlation alignment \cite{sun2016deep, chen2019joint}, and the Wasserstein metric \cite{chen2019deep, lee2019sliced}. MMD was first introduced for the two-sample tests of the hypothesis that two distributions are equal based on observed samples from the two distributions \cite{gretton2012kernel}, and this is currently the most widely used metric to measure the distance between two feature distributions. The Deep Domain Confusion Network proposed by Tzeng et al.\cite{tzeng2014deep} learns both semantically meaningful and domain invariant representations, while  Long et al. proposed DAN \cite{long2015learning} and JAN \cite{long2017deep} which both perform domain matching via multi-kernel MMD (MK-MMD) or a joint MMD (J-MMD) criteria in multiple domain-specific layers across domains. 

\textit{Contrastive Learning}: Recently, contrastive learning has achieved state-of-the-art performance in representation learning, leading to state-of-the-art results in computer vision. The aim is to learn an embedding space where positive pairs are pulled together, whilst negative pairs are pushed away from each other. Positive pairs are drawn by pairing the augmentations of the same image, whereas the negative pairs are drawn from different images. Existing contrastive learning methods have different strategies to generate positive and negative samples. Wu et al.\cite{wu2018unsupervised} maintains all the sample representations of the images in a memory bank, MoCo \cite{he2020momentum} maintains an on-the-fly momentum encoder along with a limited queue of previous samples, Tian et al.\cite{tian2019contrastive} uses all the generated multi view samples with the mini-batch approach, whereas both SimClr V1 \cite{chen2020simple} and SimClr V2 \cite{chen2020big} use momentum encoder and utilize all the generated sample representations within the mini batch. The above methods can provide a pretrained network for a downstream task, but do not consider domain shift if they are applied directly. However, our approach aims to learn representations that are generalizable without any need of labeled data. Recently, contrastive learning was applied in Unsupervised Domain Adaptation setting \cite{kang2019contrastive, park2020joint, kim2020cross}, where models have access to the source labels and/or used models pretrained on imagenet as their backbone network.  In comparison, our work is based on contrastive learning, which is also referred to as unsupervised representation learning, without having access to labeled data or pretrained imagenet parameters, but instead leveraging the vast amount of unlabeled source and target data to train a encoder from random initialized parameters. 

\textit{Removal of false negatives}: As the name suggests, contrastive learning methods learn by contrasting semantically similar and dissimilar pairs of samples. They rely on the number of negative samples for generating good quality representations and favor large batch size.  As we do not have access to labels, when an anchor image is paired with the negative samples to form a negative pair, there is a probability that these images could share the same class, in which case the contribution towards the contrastive loss becomes minimal, limiting the ability of the model to converge quickly. These false negatives remain a fundamental problem in contrastive learning methodology, but relatively limited work has been in this area thus far. 

Most existing methods focus on mining hard negatives;  \cite{kalantidis2020hard} developed hard negative mixing to synthesize hard negatives on the fly in the embedding space,  \cite{robinson2020contrastive} developed new sampling methods for selecting hard negative samples where the user can control the hardness, \cite{huynh2020boosting} proposed an approach for false negative elimination and false negative attraction and \cite{chuang2020debiased} developed a debiased contrastive objective that corrects for the sampling bias of negative samples. \cite{huynh2020boosting} use additional support views and aggregation as part of their elimination and attraction strategy. Regarding our proposed approach and inspired by \cite{huynh2020boosting}, we have further simplified and only applied the false elimination part to the domain adaptation framework. Instead of using additional support views, we compute the similarity loss between the anchor and the negatives in the mini-batch, we then sort the corresponding negative pair similarity losses for each anchor and remove the negative pairs similar to the anchor. 
For each anchor in the mini-batch, we remove the exact same number of negative pairs, for example in $FNR_1$ we remove one potential false negative from a total of 1023 negative samples with a batch size of 512, totalling 512 total potential false negatives for all the anchor images in the mini-batch of 512.

\begin{algorithm}
    \SetKwFunction{isOddNumber}{isOddNumber}
    \SetKwInOut{KwInput}{Input}
    \SetKwInOut{KwOutput}{Output}
    \KwInput{Source Data S:${(x_1^s, ...., x_n^s)}$,  \\ Target Data T:${(x_1^t,...., x_n^t)}$ }
    \KwOutput{Encoder network $f(.)$, Projection-head network $g(.)$}
    \For{sampled minibatch}{
         Make two augmentations per source image $a_1^s, a_2^s \sim S$\\
         \# source augmentation-1 \\
         $h_1^s=f(a_1^s)$ \\
         $z_1^s=g(h_1^s)$ \\
         \# source augmentation-2 \\
         $h_2^s=f(a_2^s)$ \\
         $z_2^s=g(h_2^s)$ \\
         Calculate $L_{FNR\_S}$ for $z_1^s, z_2^s$ using Eq-\ref{FNR-Loss}\\
         Make two augmentations per target image $a_1^t, a_2^t \sim T$ \\
         \# target augmentation-1 \\
         $h_1^t=f(a_1^t)$ \\
         $z_1^t=g(h_1^t)$ \\
         \# target augmentation-2 \\
         $h_2^t=f(a_2^t)$ \\
         $z_2^t=g(h_2^t)$ \\
         Calculate $L_{FNR\_T}$ for $z_1^t, z_2^t$ using Eq-\ref{FNR-Loss}\\
         Calculate $L_{FNR\_DA}$ using Eq-\ref{FNR-DA}\\
         Calculate $L_{MMD}$ using Eq-\ref{MMD loss} \\
         Update $f(.)$ and $g(.)$ by back propogating $L_{FNR\_DA}$ and $L_{MMD}$
    }
    \caption{Algorithm for Contrastive Domain Adaptation with False Negative Removal and Maximum Mean Discrepancy summarizing our proposed method.}
\end{algorithm}

\section{Method}

\subsection{Model Overview}
Contrastive Domain Adaptation (CDA):
We explore a new domain adaptation setting in a fully self-supervised fashion without any labelled data, that is where both the source and the target domain contains unlabelled data. In the normal UDA setting, we have access to the source domain whereas, our goal is to train a model using these unlabelled data sources in order to generalize visual features in both the domains. The aim is to obtain pre-trained weights that are robust to domain-shift and generalizable to the downstream domain adaptation task. Our model uses unlabelled source and target datasets in an attempt to learn and solve the adaptation between domains.

\begin{figure}[t!]
\centering
\includegraphics[width=0.4\textwidth]{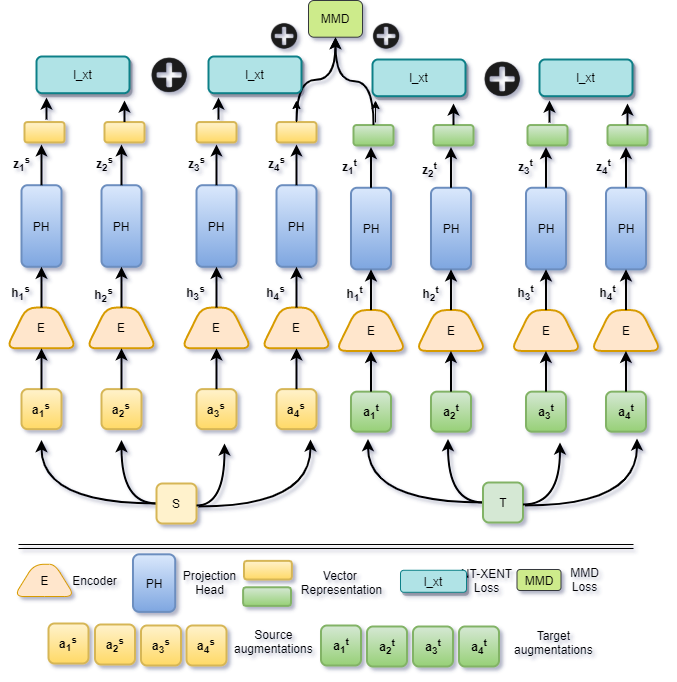}
\caption{Overview of CDA with four views}
\setlength{\belowcaptionskip}{-4pt}
\label{fig-Network-4views}
\end{figure}

Inspired by the recent successes of learning from unlabeled data, the proposed learning framework is based on SimClr \cite{chen2020simple} for domain adaptation setting where data from unlabelled source and target domains is used in task-agnostic way. SimClr \cite{chen2020simple} method learns visual similarity where a model pulls together visually similar-looking images nearby while pushing away dissimilar-looking images. However, in domain adaptation, the same class images may look very different due to domain gap, so that learning visual similarity alone does not ensure semantic similarity and domain-invariance between domains. So using CDA, we aim to learn general visual class-discriminative and domain-invariant features from both the domains via unsupervised pretraining. We introduce each specific component in detail below which is illustrated in Figure-\ref{fig-Network} and Figure-\ref{fig-Network-4views} for four views.

From randomly sampled mini-batch of images N, we augment each image S twice creating two views of same anchor image $s_i$ and $s_j$. We use base encoder(Resnet50 architecture \cite{he2016deep}) that is trained from scratch to encode augmented images in order to generate representations $hs_i$ and $hs_j$. These representations are then inputted into a non-linear MLP with two hidden layers to get the projected vector representations $zs_i$ and $zs_j$. We find that this MLP projection benefits our model by compressing our images into a latent space representation, enabling the model to learn the high-level features of the images. We apply contrastive loss on the vector representations using the NT-Xent loss \cite{chen2020simple} that has been modified to identify and eliminate false negatives, thus resulting in improved accuracy, details of which are discussed in section 4.2. We also introduce MMD to measure domain discrepancy in feature space in order to minimize domain shift, details of which are discussed later in this section. The aim is to obtain the pretrained weights that are robust to domain-shift and efficiently generalizable. In the later stage, we perform linear evaluation using the encoder whilst entirely discarding the MLP projection head after pretraining.

\subsection{Contrastive Loss for DA setting:} 
The goal of contrastive learning is to maximize the similarities between positive pairs and minimize the similarities of negative ones. We randomly sample mini batch of N images, each anchor image x is augmented twice creating two views of the same sample $x_i$ and $x_j$, resulting in 2N images. We do not explicitly sample the negative pairs, we instead follow \cite{chen2020simple}, and treat other 2(N-1) augmented image samples as negative pairs. The contrastive loss is defined as follows:
 
\begin{equation}\label{Simclr-loss}
   L_{CONT} = -log\frac{exp \left (  \\sim \left ( z_{i}, z_{j} \right)/ T \right )} {\sum_{k=1}^{2N} 1_{\left ( k\neq i \right )} \\sim \left (z_{i}, z_{k} \right )/ T} 
\end{equation}

where sim(u, v) is a cosine similarity function $u^Tv/\left \| u \right \|\left \| v \right \|$ and $T$ is a temperature parameter.

However If we use the above contrastive loss is used in a domain adaptation scenario,  as the mini-batch contains image samples from both domains, it may treat all other samples as negatives against the anchor image even though they may belong to the same class without distinguishing domains which could further widen the distance between them due to the difference in the domain specific visual characteristics, and therefore unable to learn domain invariance. In order to overcome these problems, we propose to use perform contrastive learning in the source and target domain independently by randomly sampling instances from source and target domain. Finally, our contrastive loss for DA is defined as follows:

\begin{equation}\label{contrastive-loss-da}
L_{CONT\_DA} = L_{CONT\_S} + L_{CONT\_T}
\end{equation}

where $L_{CONT\_S}$ and $L_{CONT\_T}$ are source contrastive loss and target contrastive loss

\subsection{Removal of False Negatives:} Unsupervised contrastive representation learning methods aim to learn by contrasting semantically positive and negative pairs of samples. As we do not have access to the true labels in this type of setting, positive pairs are drawn by pairing the augmentations of the same image whereas the negative pairs are drawn from different images within the same batch. So, for a batch of N images, augmented images form N positive pairs for a total of 2N images and 2N-1 negative pairs. From those 2N-1, there could be images which are similar to the anchor, hence treated as false negative.

During training, an augmented anchor image is compared against the negative samples to contribute towards a contrastive loss, as a result, there is a  possibility that some of these pairs may have the same semantic information (label) as that of the anchor, and therefore can be treated as false negatives. But in cases where the original image sample and a negative image sample share the same class, the contribution towards the contrastive loss becomes minimal, limiting the ability of the model to converge quickly,  as the presence of these false negatives discard semantic information leading to significant performance drop, we therefore identify and remove the negatives that are similar to the anchor in order to improve the performance of the contrastive learning.

Inspired by \cite{huynh2020boosting}, we have simplified and only applied the false elimination part to the domain adaptation framework in order to improve the performance of contrastive learning. Instead of using additional support views, we compute the similarity loss between the anchor and the negatives within the mini-batch, we then sort the corresponding negative pair similarity losses for each anchor and remove the negative pairs' similar to the anchor. For each anchor we remove the same number of negative pairs, example $FNR_1$ removes 1 negative pair per anchor in the mini-batch.

After removing the false negatives, the contrastive loss can be defined as follows:
 \begin{equation}
 \label{FNR-Loss}
   L_{FNR} = -log\frac{exp \left (  \\sim \left ( z_{i}, z_{j} \right)/ T \right )} {\sum_{k=1}^{2N} 1_{\left ( k\neq i,k\neq S_{i} \right )} \\sim \left (z_{i}, z_{k} \right )/ T} 
\end{equation}

where Si is the set of the negative pair that are similar to the anchor i.

However If we use the above loss is used in a domain adaptation scenario, similar to the contrastive loss, as the mini-batch contains image samples from both domains, it may treat all other samples as negatives against the anchor image even though they may belong to the same class without distinguishing domains, further widening the distance between them due to the difference in the domain specific visual characteristics, and therefore unable to learn domain invariance. In order to overcome these problems, we propose to use FNR loss in the source and target domain independently by randomly sampling instances from source and target domain. Finally, our joint FNR loss for DA is defined as follows:

\begin{equation}\label{FNR-DA}
L_{FNR\_DA} = L_{FNR\_S} + L_{FNR\_T}
\end{equation}

where $L_{FNR\_S}$ and $L_{FNR\_T}$ are source contrastive loss and target contrastive loss

\subsection{Revisiting Maximum Mean Discrepancy(MMD):}
MMD defines the distance between the two distributions with their mean embeddings in the Reproducing Kernel Hilbert Space (RKHS). MMD is a two sample kernel test to determine whether to accept or reject the null hypothesis $p = q$ \cite{gretton2012kernel}, where $p$ and $q$ are source and target domain probability distributions. MMD is motivated by the fact that if two distributions are identical, all of their statistics should be the same. The empirical estimate of the squared MMD using two datasets is computed by the following equation:



\begin{equation}\label{MMD loss}
\begin{split}
L_{{MMD}} & = \left\|\frac{1}{N}\sum_{i=1}^N\phi (x^s_i)-\frac{1}{M}\sum_{j=1}^M\phi (x^t_j) \right\|^2_{H}... \\
 & = \frac{1}{N^2}\sum_{i=1}^N\sum_{i^\prime=1}^Nk(x^s_i,x^s_{i^\prime})-\frac{2}{NM}\sum_{i=1}^N\sum_{j=1}^Mk(x^s_i,x^t_j) \\
 & + \frac{1}{M^2}\sum_{j=1}^M\sum_{j\prime=1}^Mk(x^t_j,x^t_{j^\prime}) 
\end{split}
\end{equation}


where $ \phi\left ( . \right ) $ is the mapping to the RKHS H, $k\left ( .,. \right ) = \left \langle \phi\left ( . \right ) , \phi\left ( . \right )  \right \rangle$ is the universal kernel associated with this mapping, and N, M are the total number of items in the source and target respectively. In short, the MMD between the distributions of two datasets is equivalent to the distance between the sample means in a high-dimensional feature space.




\begin{figure}[t!]
\centering
\centerline{\includegraphics[width=90mm,scale=0.5]{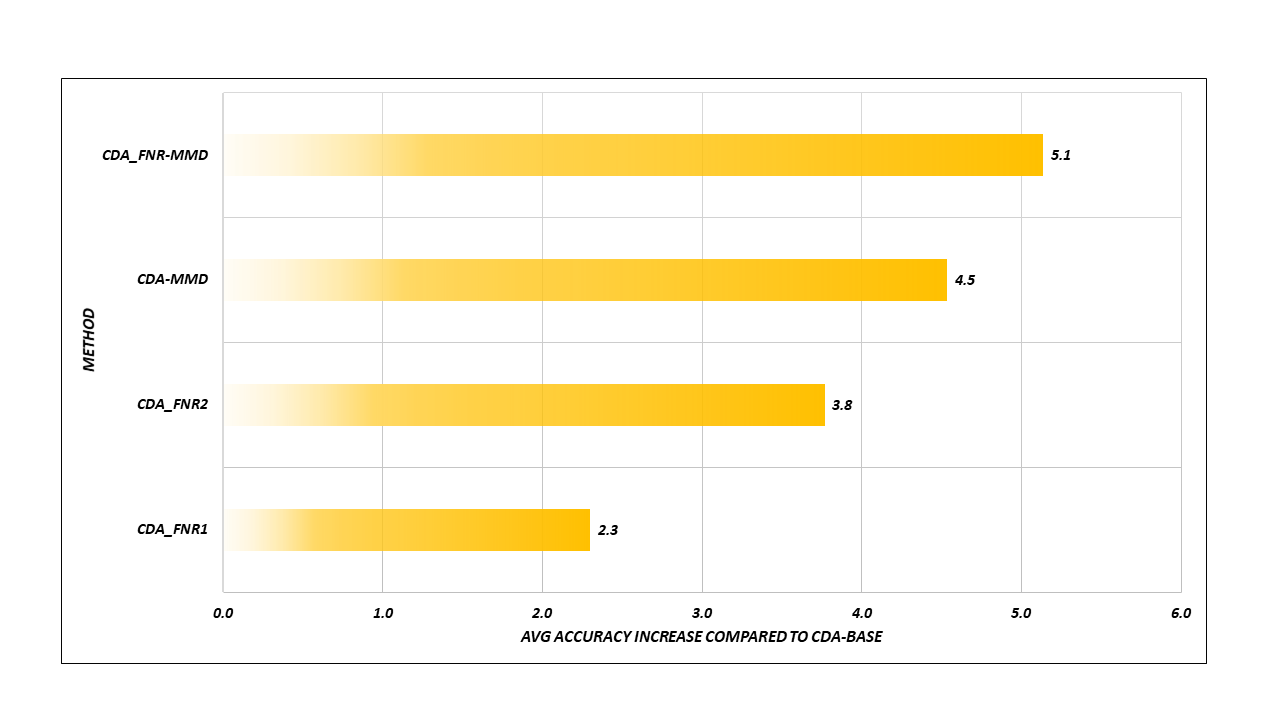}}
\caption{Average Accuracy comparision of proposed CDA frameworks with CDA-Base.}
\label{fig-FNR}
\end{figure}

\section{Experiments}
\subsection{Datasets}
Since  we  propose  a  new  task,  there  is  no benchmark that is specifically designed for our task. We illustrate the performance of our method for the contrastive domain adaptation task comparing with the SimClr-Base and CDA-Base explained later in the section-4.2, we apply it on the standard digits dataset benchmarks using accuracy as the evaluation metric.

MNIST $\longrightarrow$ USPS (M$\rightarrow$U): MNIST \cite{lecun1998gradient} which stands for “modified NIST” is treated as source domain, it consists of black-and-white handwritten digits and USPS \cite{denker1989neural} is treated as target domain, it consists handwritten digit datasets scanned and segmented by the U.S. Postal Service. As both these datasets contain grayscale images the domain shift between these two datasets is relatively small. Figure \ref{fig-FNR-digit1} below shows sample images from M$\rightarrow$U.

\begin{figure}[h!]
\centering
\caption{Sample images from datasets: MNIST-USPS}
\centerline{\includegraphics[width=75mm,scale=0.5]{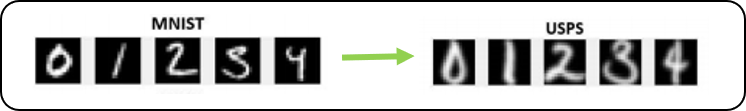}}
\label{fig-FNR-digit1}

\end{figure}

\begin{figure}[h!]
\centering
\caption{Sample images from datasets: SVHN-MNIST}
\centerline{\includegraphics[width=75mm,scale=0.5]{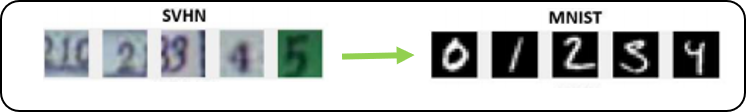}}
\label{fig-FNR-digit2}
\end{figure}

SVHN $\longrightarrow$ MNIST (M$\rightarrow$S): In this setting, SVHN \cite{netzer2011reading} is treated as source domain and MNIST \cite{lecun1998gradient} is treated as the target domain. MNIST consists of black-and-white handwritten digits, SVHN consists of crops of coloured, streetview house numbers consisting of single digits extracted from images of urban house numbers from Google Street View. SVHN and MNIST are two digit classification datasets with a drastic distributional shift between the two of them.
The adaptation from MNIST to SVHN is quite challenging because MNIST has a significantly lower intrinsic dimensionality than SVHN. Figure below \ref{fig-FNR-digit2} shows sample images from M$\rightarrow$S. 

\begin{figure}[h!]
\centering
\caption{Sample images from datasets: MNIST-MNISTM}
\centerline{\includegraphics[width=75mm,scale=0.5]{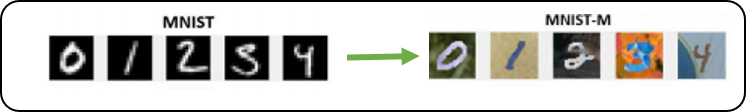}}
\label{fig-FNR-digit3}
\end{figure}

MNIST $\longrightarrow$ MNISTM (M$\rightarrow$MM): MNIST \cite{lecun1998gradient}, which consists of black-and-white handwritten digits is treated as the source domain and MNISTM is treated as a target domain. MNISTM is a modification of MNIST dataset where the digits are blended with random patches from BSDS500 dataset color photos. Figure below \ref{fig-FNR-digit3} shows sample images from M$\rightarrow$MM.

\subsection{Implementation Details}
CDA uses a base encoder ResNet-50 \cite{he2016deep} trained from scratch followed by a two layered non-linear MLP. During pretraining, we train CDA on 2 Titan Xp GPUs, using LARS optimizer \cite{you2017large} with a batch size of 512 and weight decay of le-6 for a total of 300 epochs. Similar to SimClr\cite{chen2020simple}, we report performance by training a linear classifier on top of a fixed representation, but only with source labels to evaluate representations which is a standard benchmark that has been adopted by many papers in the literature \cite{chen2020simple,chen2020big,oord2018representation}.

\subsection{Evaluation}
We conducted various experiments using unlabeled source and target digit datasets. The goal of our experiments is to introduce contrastive learning to the domain adaptation problem in order to maximize generalization between source and target datasets by learning class discriminative and domain-invariant features along with improving the performance of contrastive loss by eliminating the false negatives which is one of the main drawbacks in the contrastive learning without access to labels. We have performed multiple experimented using two views and four views \cite{tian2019contrastive}. 
Figure-\ref{fig-FNR} compares the average accuracy of our proposed two view CDA frameworks with the CDA-Base.
Following are the various experimental scenarios we performed on the digit datasets.

SimClr-Base: We start our experimental analysis by evaluating using SimClr. We have trained on source dataset using the same setup as SimClr, whilst testing on the target dataset. We treat this as a strong baseline which we call SimClr-Base and use this as reference for comparison against other methods. 

CDA-Base: We followed the methodology as described in section-3.2, trained the model using the equation-\ref{contrastive-loss-da} and evaluated on the target domain. Looking at table-\ref{tab:FNR-Removal-Table}, we can clearly observe that the model shows higher performance compared to the SimClr-Base. The model has clearly learnt both visual similarity and domain-invariance resulting in minimizing the distance between the domains and maximizing the classification accuracy. Overall, the average accuracy for all the datasets has increased by around 19\% compared to the SimClr-Base model. We treat this result as a second strong baseline and call it CDA-Base.

CDA\_FNR: We followed the methodology as described in section-3.3, and trained the model using the equation-\ref{FNR-DA}. We then evaluated the model trained using this method on the target domain. Looking at table-\ref{tab:FNR-Removal-Table}, in addition to learning visual similarity and domain-invariance, model also successfully identified and eliminated the potential false negatives as they contain the same semantic information as that of the anchor, resulting in converging faster and increased accuracy. We experimented on two scenarios, firstly we removed one false negative which we call FNR1 and in the second case, we experimented by removing two false negatives which we call FNR2.  The results of these experiments can be seen in table-\ref{tab:FNR-Removal-Table}, which concludes that removal of false negatives improves accuracy resulting in converging faster. The average accuracy has increased by 2.3\% after removing one false negative. Additionally by removing two false negatives, we observe that the average accuracy has increased by 3.8\% in comparison to CDA-Base and 1.5\% in comparison to FNR1. Compared to the SimClr-Base, the average accuracy has increased around 21\%.

CDA-MMD: We have used the same setup as that of CDA-Base. Additionally we introduced MMD as described in section 3.4, which is computed between vector representations extracted from each domain as per the equation-\ref{MMD loss}, in order to reduce the distance between the source and target distributions. We backpropagate NT-Xent loss from equation-\ref{contrastive-loss-da} along with MMD loss equation-\ref{MMD loss}. From table-\ref{tab:my-table-MMD}, we observe that by minimizing both these losses together, our model achieves much better alignment of the source and target domains, showing the usefulness of combined contrastive loss and MMD alignment. In comparison to the CDA-Base method, the performance gain tends to be large as we can see that it has increased by 4.5\%. 

CDA\_FNR-MMD: We have used the same setup as that of CDA\_FNR, additionally we have introduced MMD which is computed between vector representations extracted from each domain as per the equation-\ref{MMD loss}, in order to reduce the distance between the source and target distributions. We calculate FNR loss both for source and target domains using equation-\ref{FNR-DA} and backpropagate FNR loss equation-\ref{FNR-DA} along with MMD loss equation-\ref{MMD loss}. From table-\ref{tab:my-table-MMD}, we observe that by removing the potential false negatives and minimizing the discrepancy together, our model retains semantic information, hence converges faster and learns both visual similarity and domain-invariance by aligning source and target datasets efficiently, showing the effectiveness of this method. In comparison to the CDA-Base method, the average performance gain tends to be larger as we can see that it has increased by huge margin of 5.1\%.  

Comparison with the state of art methods: As the proposed framework is new, unfortunately, there are no benchmarks specifically designed for our task so it’s very difficult for a like for like comparison. Using our method, we demonstrate that our model can perform as well to the domain adaptation setting without access to labelled data and imagenet parameters, just by training using the unlabeled data itself, whereas all the unsupervised domain adaptation methods have access to the source labels. We have compared our results with the state-of-the-art models and can conclude our model performs favorably in comparison with other state-of-the art models. From table-\ref{tab:my-table-other-methods}, we can conclude that our model has outperformed in the MNIST-USPS and SVHN-MNIST tasks compared to the other state-of-the-art models like DANN, DAN, ADDA, DDC and Simclr-Base \cite{ganin2016domain, long2015learning, tzeng2017adversarial, liu2016coupled, tzeng2014deep, chen2020simple}

\begin{table}[]
\centering
\caption{Accuracy values on the digits datasets evaluated using the proposed SimClr-Base and proposed CDA framework, along with the introduction of false negative removal. The best average is indicated in \textbf{bold}. M:MNIST, U:USPS, S:SVHN and MM:MNISTM.}
\label{tab:FNR-Removal-Table}
\begin{tabular}{lcccc}
\hline
 Method & \multicolumn{1}{l}{\begin{tabular}[c]{@{}l@{}}M$\rightarrow$U\end{tabular}} & \multicolumn{1}{l}{\begin{tabular}[c]{@{}l@{}}S$\rightarrow$M\end{tabular}} & \multicolumn{1}{l}{\begin{tabular}[c]{@{}l@{}}M$\rightarrow$MM\end{tabular}} & Avg  \\ \hline
SimClr-Base \cite{chen2020simple} & 92.0 & 31.7 & 34.9 & 53.1   \\
CDA-Base & 92.5 & 64.8 & 57.9 & 71.7  \\ 
CDA\_FNR1 & 93.2 & 69.4 & 59.5 & 74.0   \\ 
CDA\_FNR2 & 94.1 & 71.7 & 60.6 & \textbf{75.5} \\ \hline
\end{tabular}
\end{table}

Inspired by \cite{tian2019contrastive}, we have also performed similar experiments using four views, following are the experimental scenarios we performed on the digit datasets which we compare with the CDA-Base and Contrastive Domain Adaptation with Four Augmentations(CDAx4aug).

CDAx4aug: We have tested our method by using four augmentations per anchor per source and followed the methodology as described in section-3.2, by training the model using the equation-\ref{contrastive-loss-da}. The only change is that we now backpropagate four contrastive losses two from source and two from target. From table-\ref{tab:Multiview-FNR-Removal-Table}, we can observe that the additional augmentations have significantly improved the average method accuracy compared to the two view CDA-Base due to the availability of additional positive and negative samples. Overall, by adding two additional views to the CDA-Base method we have gained an average accuracy of 5.1\% compared to the CDA-Base method.

\begin{table}[]
\centering
\caption{Accuracy values on the digits datasets evaluated using CDA framework with the introduction of MMD and compared against base models. The best average is indicated in \textbf{bold}. M:MNIST, U:USPS, S:SVHN and MM:MNISTM.}
\label{tab:my-table-MMD}
\begin{tabular}{lcccl}
\hline
Method & \multicolumn{1}{l}{\begin{tabular}[c]{@{}l@{}}M$\rightarrow$U\end{tabular}} & \multicolumn{1}{l}{\begin{tabular}[c]{@{}l@{}}S$\rightarrow$M\end{tabular}} & \multicolumn{1}{l}{\begin{tabular}[c]{@{}l@{}}M$\rightarrow$MM\end{tabular}} & Avg \\ \hline
SimClr-Base \cite{chen2020simple} & 92.0 & 31.7 & 34.9 & 53.1 \\
CDA-Base & 92.5 & 64.8 & 57.9 & 71.7 \\ 
CDA-MMD & 93.4  & 74.8 & 60.6 & 76.2 \\
CDA\_FNR-MMD & 94.2 & 76.2 & 60.2 & \textbf{76.8}  \\ \hline
\end{tabular}
\end{table}

\begin{table}[]
\centering
\caption{Comparision of the proposed CDA method with state-of-the-art methods, using ACCURACY as the performance metric. The best numbers are indicated in \textbf{bold}. M:MNIST, U:USPS, S:SVHN and MM:MNISTM.}
\label{tab:my-table-other-methods}
\begin{tabular}{lccc}
\hline
Method & \multicolumn{1}{l}{\begin{tabular}[c]{@{}l@{}}M$\rightarrow$S\end{tabular}} & \multicolumn{1}{l}{\begin{tabular}[c]{@{}l@{}}S$\rightarrow$M\end{tabular}} & \multicolumn{1}{l}{\begin{tabular}[c]{@{}l@{}}M$\rightarrow$MM\end{tabular}} \\ \hline
SimClr-Base \cite{chen2020simple} & 92.0 & 31.7 & 34.9 \\ 
DDC & 79.1  & 68.1 & - \\
ADDA & 89.4 & 76.0 & - \\ 
DANN & - & 73.8 & 76.6 \\ 
DAN & 81.1 & 71.1 & \textbf{76.9} \\ 
CDA\_FNR-MMD\\ (our method) & \textbf{94.2} & \textbf{76.2} & 60.2 \\ \hline
\end{tabular}
\end{table}

\begin{table}[]
\centering
\caption{Accuracy values on the digits datasets compared with Base models and evaluated using CDA framework with four views along with the introduction of false negative removal. The best average is indicated in \textbf{bold}. M:MNIST, U:USPS, S:SVHN and MM:MNISTM.}
\label{tab:Multiview-FNR-Removal-Table}
\begin{tabular}{lcccc}
\hline
Method & \multicolumn{1}{l}{\begin{tabular}[c]{@{}l@{}}M$\rightarrow$U\end{tabular}} & \multicolumn{1}{l}{\begin{tabular}[c]{@{}l@{}}S$\rightarrow$M\end{tabular}} & \multicolumn{1}{l}{\begin{tabular}[c]{@{}l@{}}M$\rightarrow$MM\end{tabular}} & \textbf{Avg} \\ \hline
SimClr-Base \cite{chen2020simple} & 92.0 & 31.7 & 34.9 & 53.1 \\
CDA-Base & 92.5 & 64.8 & 57.9 & 71.7 \\ 
CDAx4aug & 92.9 & 74.1 & 63.5 & 76.8 \\ 
CDAx4aug\_FNR & 93.6 & 75.0 & 64.0 & \textbf{77.5}  \\ \hline
\end{tabular}

\end{table}

\begin{table}[]
\centering
\caption{Accuracy values on the digits datasets evaluated using CDA framework with four views, along with the introduction of MMD compared with Base models. The best average is indicated in \textbf{bold}. M:MNIST, U:USPS, S:SVHN and MM:MNISTM.}
\label{tab:Introducing MMDFNR multiple views}
\begin{tabular}{lcccc}
\hline
Method &
  \begin{tabular}[c]{@{}c@{}}M$\rightarrow$S\end{tabular} &
  \begin{tabular}[c]{@{}c@{}}S$\rightarrow$M\end{tabular} &
  \begin{tabular}[c]{@{}c@{}}M$\rightarrow$MM\end{tabular} &
  Avg \\ \hline
SimClr-Base \cite{chen2020simple} & 92.0 & 31.7 & 34.9 & 53.1 \\
CDA-Base & 92.5 & 64.8 & 57.9 & 71.7 \\ 
CDAx4aug & 92.9 & 74.1 & 63.5 & \textbf{76.8} \\ 
CDAx4aug-MMD & 92.7 & 69.3 &  58.6 & 73.5 \\ 
CDAx4aug\\\_FNR-MMD &  92.5    &  70.6    &  61.5    &   74.9   \\ \hline
\end{tabular}
\end{table}



CDAx4aug\_FNR: We followed the methodology as described in section-3.3, and trained the model using the equation-\ref{FNR-DA}, by training the model on four augmentations per domain as opposed to two. We then evaluated the model trained using this method on the target domain. Looking at table-\ref{tab:Multiview-FNR-Removal-Table}, we can clearly establish that the additional views helped the model learn visual similarity and domain-invariance resulting in minimizing the distance between the domains, it also helped the model with successful identification and elimination of the potential false negatives, thus resulting in converging faster with average accuracy increase of 5.8\% compared to CDA-Base and 0.7\% compared to CDAx4aug-Base. 

CDAx4aug-MMD: We have used the same setup as that of CDAx4aug, additionally we introduced MMD computed between vector representations extracted from each domain as per the equation-\ref{MMD loss}. We backpropagate XT-Xent loss equation-\ref{contrastive-loss-da} for two pairs of source and two pairs of target along with MMD loss-\ref{MMD loss}. From table-\ref{tab:Introducing MMDFNR multiple views} we can observe that performance gain using MMD was not significant due to the noise from additional augmentations, resulting in slow convergence between the source and target distributions. 

CDAx4aug\_FNR-MMD: We have used the same setup as that of CDAx4aug\_FNR, additionally we have introduced MMD computed between vector representations extracted from each domain as per the equation-\ref{MMD loss}. We backpropagate FNR loss using equation-\ref{FNR-DA} along with MMD loss using equation-\ref{MMD loss}. From table-\ref{tab:Introducing MMDFNR multiple views}, we can see that the average performance gain has increased compared to CDAx4aug-MMD method due to the false negative removal, but addition of MMD has comparatively slowed the convergence.

\section{Conclusion}
Over the past few years, ImageNet pre-training has become a standard practice. Employing our proposed contrastive domain adaptation approach and its variants, we demonstrate that our model can perform competitively in a domain adaptation setting, without having access to labelled data or imagenet parameters, just by training using the unlabeled data itself. CDA also introduces identification and removal of the potential false negatives in the DA setting, resulting in improved accuracy. We also extend our framework to learn from more than two views in the DA setting. We tested our model using various experimental scenarios demonstrating that it can be effectively used for downstream domain adaptation task. We hope that our work encourages future researchers to apply contrastive learning to domain adaptation.

{\small
\bibliographystyle{ieee}

}

\end{document}